\documentclass{article}

\usepackage{arxiv}

\usepackage[utf8]{inputenc} 
\usepackage[T1]{fontenc}    
\usepackage{hyperref}       
\usepackage{url}            
\usepackage{booktabs}       
\usepackage{amsfonts}       
\usepackage{nicefrac}       
\usepackage{microtype}      
\usepackage{lipsum}

\usepackage{graphicx}

\title{Evolving Self-taught Neural Networks:\\The Baldwin Effect and the Emergence of Intelligence}

\author{
  Nam Le\thanks{Independent Researcher, Hanoi, Vietnam, email: namlehai90@gmail.com}
        \thanks{Natural Computing Research \& Applications,
        University College Dublin,
        Dublin, Ireland,
        email: nam.lehai@ucdconnect.ie}
}

\begin{document}
\maketitle

\footnote{Preprint AISB’19, 16th April 2019, AISB 2019 Symposium}\begin{abstract}
The so-called \textbf{Baldwin Effect} generally says how learning, as a form of \textit{ontogenetic} adaptation, can influence the process of \textit{phylogenetic} adaptation, or evolution. This idea has also been taken into computation in which evolution and learning are used as computational metaphors, including evolving neural networks. This paper presents a technique called evolving \textit{self-taught} neural networks -- neural networks that can teach themselves \textit{without external supervision or reward}. The self-taught neural network is \textit{intrinsically motivated}. Moreover, the self-taught neural network is the product of the interplay between evolution and learning. We simulate a multi-agent system in which neural networks are used to control autonomous agents. These agents have to forage for resources and compete for their own survival. Experimental results show that the interaction between evolution and the ability to teach oneself in self-taught neural networks outperform evolution and self-teaching alone. More specifically, the emergence of an intelligent foraging strategy is also demonstrated through that interaction. Indications for future work on evolving neural networks are also presented.
\end{abstract}


\section{Introduction}

Evolution and learning are two forms of adaptation. The former is a change at \textit{genotypic} level of a population, also called \textit{phylogenetic} adaptation. The latter is a change at \textit{phenotypic} level of an individual as a result of experience with its environment during lifetime. Thus, learning is a form of lifetime or \textit{ontogenetic} adaptation. Reasonably, lifetime adaptation takes place at a quicker pace than evolution, preparing the organism for increasingly uncertain environments which may require some survival skill that the slower evolutionary process could not fully offer.

Interestingly, evolution and learning can complement each other through the phenomenon called the Baldwin Effect \cite{Baldwin:1896}, which was first demonstrated computationally by Hinton and Nowlan (henceforth H\&N) \cite{Hinton:1987}. Following this success, there have been quite a few important studies studying the interaction between learning and evolution, notably in evolving neural networks \cite{Nolfi:1994}, \cite{Parisi91learning}, and in the NK-Landscape \cite{Mayley:1997}. Regardless of the problem domain and how learning is implemented, most studies focus on how learning and evolution are combined to solve an \textbf{individual problem} in a sort-of \textbf{single-agent} environment. This means each agent has its own problem (though they are copies of each other) to solve. There is no interactive effect between the agents and their solutions to each other. This differs greatly in the case of a multi-agent environment in which agents live in the same environment and may have to compete and cooperate in solving their own problems or problems shared with others.

Although there should possibly be a mixture of flavour, this paper aims at two main things. First, we present a technique called evolving self-taught neural networks, or neural networks that can teach themselves without external supervisory signals. This is an important aspect of this contribution. Second, we simulate a multi-agent foraging world to test the performance of our proposed method and see the effect of interest. More specifically, we shall be seeing how evolution and the ability of self-teaching interact with each other in creating more adaptive and autonomous foraging agents, those that have little knowledge about the world.

In the remainder of this paper, we initially present some prior research relating to the Baldwin Effect, including some review on learning and evolution in neural networks. We then describe the detail of the neural network and the simulation undertaken. The results from these experiments are analysed and discussed, and finally, conclusions and several interesting future research opportunities are proposed.

\section{Related Work}

\subsection{The Baldwin Effect}

In nature, the organism with learning ability may be able to learn some new skill or knowledge to adapt as the environment becomes harder or unpredictable that what evolution has provided is not sufficient to survive. The Baldwin Effect is often understood as, over generations, that skill or knowledge becomes innate or closer to be innate so that the future organism can quickly adapt to the environment with fewer or even without any learning effort undertaken \cite{Simpson:1953}. This shows how learning, or lifetime adaptation, can influence the evolutionary pathway of a species. 

The idea that learning can influence evolution in Darwinian framework was discussed by psychologists and evolutionary biologists over one hundred years ago through \textit{`A new factor in evolution'} \cite{Baldwin:1896}, \cite{Morgan:1896}, \cite{Simpson:1953}. However, it gradually gained more attention since the classic paper in 1987 by the British Cognitive Scientist Geoffrey Hinton and his colleague Steven Nowlan at CMU (\cite{Hinton:1987}). Hinton and Nowlan (henceforth H\&N) demonstrated an instance of the Baldwin effect in a computer simulation. They used a Genetic Algorithm to evolve a population in a Needle-in-a-haystack landscape showing that learning can help evolution to search for a solution when evolutionary search alone is ineffective. Through the Baldwin-like effect in H\&N's simulation, the correct behaviour (solution) can gradually emerge by the interaction between learning and evolution, but cannot happen by both learning or evolution alone \cite{Hinton:1987}.

The model developed by Hinton and Nowlan, though simple, is interesting, opening up the trend followed by a number of research papers investigating the interaction between learning and evolution. Following the framework of Hinton and Nowlan, there have been a number of other papers studying the Baldwin effect in the NK-fitness landscape, which was developed by Stuart Kauffman \cite{Kauffman1989} to model `tunably rugged` fitness landscapes. Problems within that kind of landscape are shown to fall in NP-completeness category \cite{Kauffman1989}. Several notable studies of the Baldwin effect in the NK-model include work by Giles Mayley \cite{Mayley:1997} in which the Baldwin Effect was shown to occur as learning can guide evolution to cope with the rugged fitness landscape.

\subsection{Learning and Evolution in Neural Networks}
Following the work of Hinton and Nowlan \cite{Hinton:1987}, There have also been several studies on the topic of learning and evolution in Neural Networks. Notable studies include  \cite{Keesing1990EvolutionAL} in which the authors used a genetic algorithm to evolve the initial weights of a digit classifier neural network which then can be learned by backpropagation. They found that if the amount of learning is used properly, learning can take advantage of starting weights produced by evolution to further the classification performance.

Todd and Miller \cite{Todd:1991:EAA:116517.116552} proposed an imaginary underwater environment in which each agent in one of the two feeding patches, and has to decide whether to consume substances floating by, without any feedback given to an individual agent that could be used to discriminate between food and poison. Each agent uses its neural network to associate the colour (red or green) and the substance (food or poision). Hebbian learning \cite{Hebb1949} in combination with evolution was shown to do better than both evolution and learning alone in this scenario.

Nolfi and his colleagues made a simulation of \textit{animats}, or robots, controlled by neural networks situated in a grid-world, with discrete state and action spaces \cite{Nolfi:1994}. Each agent lives in its own copy of the world, hence no mutual interaction. The evolutionary task is to evolve action strategies to collect food effectively, while each agent learns to predict the sensory inputs to neural networks for each time step. Learning was implemented using backpropagation based on the error between the actual and the predicted sensory inputs to update the weights of a neural network. It was shown that learning to predict can enhance the evolutionary search, hence increasing the performance of the robot.

Generally learning in neural networks can be thought of as part of neural plasticity. There have been some other ideas, like evolving local learning rules to update the weights \cite{Bengio-rule-learning}, evolution of neuromodulation which facilitates the information transfer between neurons in hopes of creating meta-learning \cite{DOYA-meta-neuromodulation}. Please refer to \cite{Soltoggio2018} for more recent studies on evolving plastic neural networks. In short, most of the work use \textit{disembodied} and \textit{unsituated} neural networks in single-agent environment, having no mutual interaction as they solve their own problems, having no effect on other's performance.

In this paper, we propose a neural architecture called self-taught neural networks -- neural networks that can teach themselves without an external \textit{teacher} or \textit{reward}. This differs greatly from traditional supervised learning in which a learning machine is provided with labels served as the external teacher. This technique also differs from reinforcement learning in which a learning agent has a reward provided by its external environment. Indeed, the agent controlled by the self-taught network can perform learning on its own without external reward. This type of network can be considered sort-of \textbf{intrinsically motivated}, hence the agent controlled by the network. Moreover, the self-taught network can both learn (self-teaching) and evolve. We shall be seeing how learning and evolution interact with each other in producing self-taught neural networks in later sections.

Second, we simulate a situated multi-agent system -- a system containing multiple situated agents living together and doing their tasks while competing with each other. Each agent is controlled by a neural network but situated (and has a \textit{soft-embodiment}). This means, the way an agent acts and moves in the world affects the subsequent sensory inputs, hence the future behaviour of that agent. Our simulations are described in the following section.

\section{Simulation setup}
This section describes the detail of the simulated world containing foods and agents as well as the neural network architecture used to control the agent moving and foraging in the world.

\subsection{The Simulated World of Foods and Agents}

\begin{figure}[t]
\begin{center}
\includegraphics[width=4in]{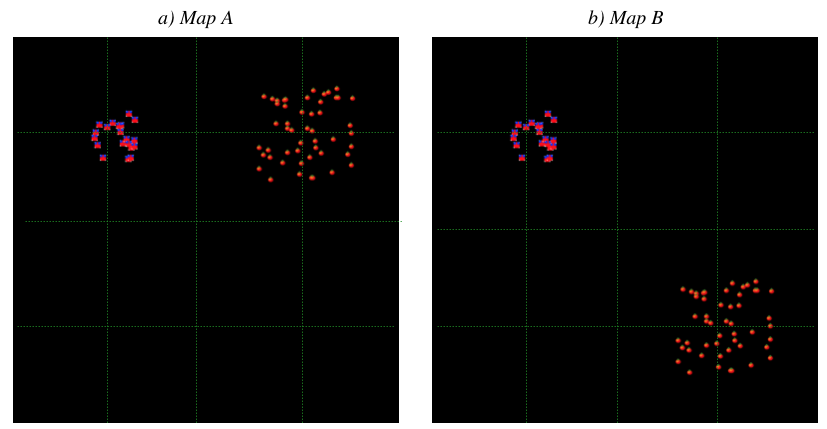}
\caption{MiniWorld -- The environment of agents and food, 640x640.}
\label{fig:env}
\end{center}
\end{figure}

Suppose that 20 agents situate in a continuous 640x640 2D-world, called \textbf{MiniWorld}, and they have to find resources to feed themselves to survive. There are 50 food particles in the world. Each food particle is represented by a square image with size 10x10. Each agent in MiniWorld also has a squared body of size 10x10. Agents and foods are initially located at separate regions in MiniWorld depending on the world map. We use two world maps (map A, and map B) in our simulations as described as follows:

Let's denote $width$ and $height$ the width and the height of MiniWorld. Initially, in both maps, all agents are located around the vicinity of radius 40 (4 times the size of an agent) around the point ($width$/4, $height$/4) (the central point of the top left quarter, as shown in Figure \ref{fig:env}).

Foods in map A have horizontal and vertical dimensions randomly chosen in ranges ($width*5/8$, $width*7/8$) and ($height*1/8$, $height*3/8$), respectively. The food region in map A is the square that has the same central point as the top right quarter, and each side of that square has the length of $width/4$. In map B, the food has its horizontal and vertical dimensions randomly chosen in ranges ($width*5/8$, $width*7/8$) and ($height*5/8$, $height*7/8$), accordingly. The food region in map B is the square that has the same center as the bottom right quarter, and each side of that square has the length of $width/4$. Two world maps are visualised in Figure \ref{fig:env} (Please note that dim green lines are sketched only for the purpose of visualising the world map of foods and agents, MiniWorld is a continuous world, not grid-like). Through the visualisation, it can be temporarily seen that map B is likely to be more difficult than map A since the food source is further to reach.

Initially agents is located far from the food source so that they have to forage to find the food source, to feed themselves. When an agent's body happens to collide with a food particle, the food particle is eaten, the energy level of the agent increases by 1, and another food piece randomly spawn in the \textbf{same region} but at a different location. The collision detection criterion is specified by the distance between the two bodies (of the agent and of the food). The agent body somehow affects how the agent senses and acts in MiniWorld. By the re-appearance of food, the environment changes as an agent eats a food.

One property of MiniWorld is it has no strict boundary, and we implement the so-called \textit{toroidal} -- this means when an agent moves beyond an edge, it appears in the opposite edge.

Each agent has a heading (in principle) of movement in the environment. Rather than initialising all agents with random headings, to make it more controllable, all the agents are initialised with a horizontal heading (i.e. with 0 degree). This somewhat explains the purpose of the design of map A and map B. In map A, all agents are initially born with a tendency to move forwards the food source. On the contrary, the agent in map B is born with a wrong direction to the food source. This clearly shows that map B is more difficult than map A. Agents in map B should have to acquire correct foraging behaviour to find the food source first, not to say they have to compete with each other for the energy. This is to say, agents in map B should develop a form of intelligent foraging to effectively seek for resources.

In our simulation, we assume that every agent has a \textit{priori} ability to sense the angle between its current heading and the food if appearing in its visual range. The visual range of each agent is a circle with radius 4. Each agent takes as inputs three sensory information, which can be the binary value 0 or 1, about what it sees from the left, front, and right in its visual range. If there is no food appearing in its visual range, the sensory inputs are all set to 0. If there is food appearing on the left (front, or right), the left (front, or right) sensor is set to 1; otherwise, the sensor is 0.

Let $\theta$ (in degree) be the angle between the agent and the food particle in its visual sense. An agent determines whether a food appears in its left, front, or right location in its visual range be the following rule: Right if 15 $<$ $\theta$ $<$ 45; Front if $\theta < 15$  or $\theta > 345$; and Left if $315 < \theta <345$. 
    
\begin{figure}[t]
\begin{center}
\includegraphics[width=4in]{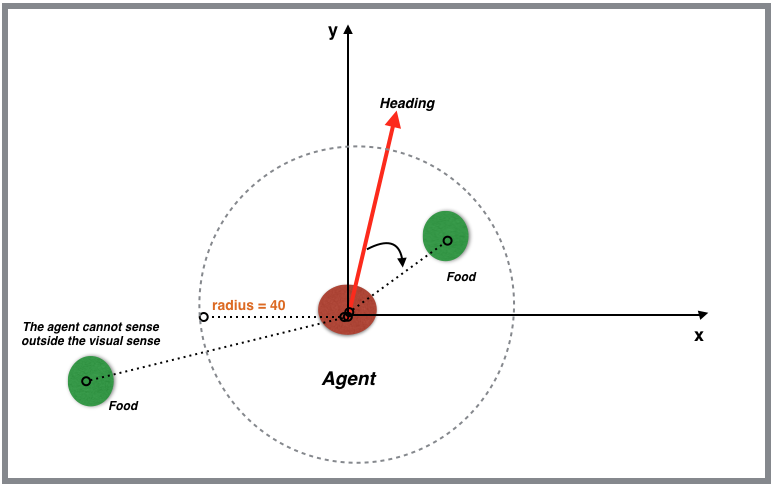}
\caption{Agent situated in an environment seeing food}
\label{fig:agent-food}
\end{center}
\end{figure}

We let all agents live in the same MiniWorld. They feed for their own survival during their life. The more an agent eats, the less the chance for others to feed themselves. This creates a stronger competition in the population.  When an agent moves for foraging, it changes the environment in which other agents live, changing how others sense the world as well. This forms a more complex dynamics, even in simple scenario we are investigating in this paper.

The default velocity (or speed) for each agent is 1. Every agent has three basic movements: Turn left by 9 degrees and move, move forward by double speed, turn right by 9 degrees and move. For simplicity, these rules are pre-defined by the system designer of MiniWorld. We can imagine the perfect scenario like if an agent sees a food in front, it doubles the speed and move forward to catch the food. If the agent sees the food on the left (right), it would like to turn to the left (right) and move forward to the food particle. The motor action of an agent is guided by its neural network as described below.

\subsection{The neural network controller}
\label{sec:ann-controller}
Each agent is controlled by a fully-connected neural network to determine its movements in the environment. What an agent decides to do changes the world the agent lives in, changing the next sensory information it receives, hence the next behaviour. This forms a sensory-motor dynamics and a neural network acts as a situated cognitive module having the role to guide an agent to behave adaptively, or \textit{Situated Cognition} even in such a simple case like what is presenting in this paper. Each neural network includes 3 layers with 3 input nodes in input layer, 10 nodes in hidden layer, and 3 nodes in output layers.

\begin{figure}[t]
\begin{center}
\includegraphics[width=4in]{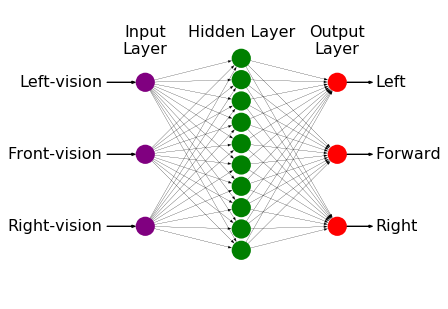}
\caption{Neural network controller for each situated agent. Connection weights can be created by evolutionary process, but can also be changed during the lifetime of an agent.}
\label{fig:neural-controller}
\end{center}
\end{figure}

The first layer takes as input what an agent senses from the environment in its visual range (described above). The output layer produces three values as a motor-guidance for how an agent should behave in the world after processing sensory information. The maximum value amongst these three values is chosen as a motor action as whether an agent should turn left, right, or move forward (as described above). All neurons except the inputs use a sigmoidal activation function. All connections (or synaptic strengths) are initialised as Gaussian(0, 1). These weights are first initialised as \textit{innate}, or merely specified by the genotype of an agent, but also have the potential to change during the lifetime of that agent.\\ 

The architectural design of neural network controller is visualised in Figure \ref{fig:neural-controller}. In fact, the neural architecture as shown in Figure \ref{fig:neural-controller} has no ability to learn, or to teach itself. In the following section, we extend this architecture to allow for self-taught learning agents.

\subsection{The Self-taught neural architecture}
\label{sec:self-taught-ann}
\begin{figure}[t]
\begin{center}
\includegraphics[width=4in]{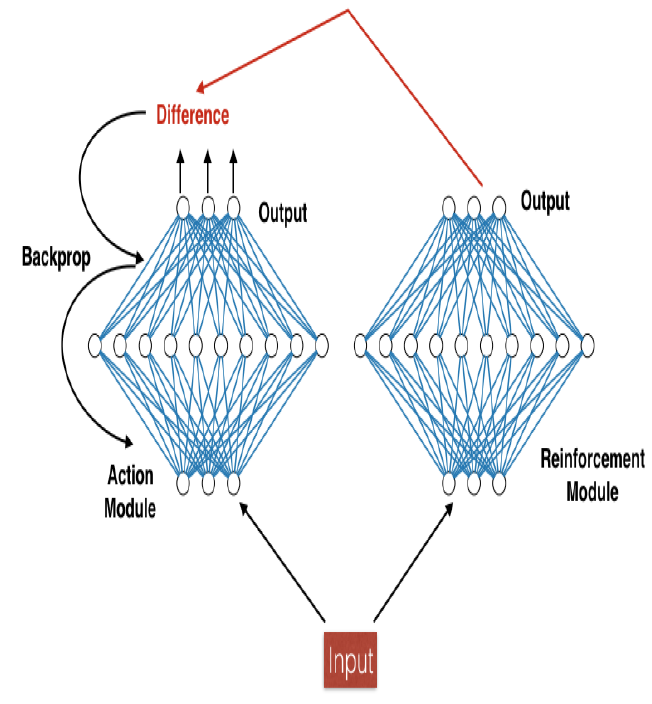}
\caption{Self-taught neural network.}
\label{fig:self-taught}
\end{center}
\end{figure}

To allow for self-taught ability, the neural controller for each agent now has two modules: one is called \textbf{Action Module}, the other is called \textbf{Reinforcement Module}. The action module has the same network as previously shown in Figure \ref{fig:neural-controller}. This module takes as inputs the sensory information and produces reinforcement outputs in order to guide the motor action of an agent. The reinforcement module has the same set of inputs as the action module, but possesses separate sets of hidden and output neurons. The goal of reinforcement network is to provide \textit{reinforcement signals} to guide the behaviour of each agent. The topology of a neural network in this case is visualised in Figure \ref{fig:self-taught}.

The difference between the output of the reinforcement module and the action module is used as the error of the output behaviour of the action module. That error is used to update the weights in action modules through Backpropagation \cite{Rumelhart1986}. Through that learning process, the action module approximates its output activation towards the output of the reinforcement module. In fact, the reinforcement and the action modules are not necessary to have the same topology. For convenience, in our simulation we allow the reinforcement module possesses the same neuronal structure as the action module, but has 10 hidden neurons separate from the hidden neurons of the action module, hence the connections. The learning rate is 0.01.

In the following sections, we describe simulations we use to investigate the evolutionary consequence of lifetime learning.

\subsection{Simulation 1: Evolution alone (EVO)}
In this simulation, we evolving a population of agents without learning ability. The neural network controller for each agent is the one described in Figure \ref{fig:neural-controller}. The \textit{genotype} of each agent is the weight matrix of its neural network, and the evolutionary process takes place as we evolve a population of weights, a common approach in Neuroevolution (NE) \cite{XinYao1999}.

Selection chooses individuals based on the number of food particles consumed. The higher the number of particles eaten, the higher the agent's fitness value.  For crossover, two individuals are selected to produce one offspring. We implement crossover as follows. The more  successful a parent, the greater the likelihood that its weights are copied to the child. Each weight element in the matrix of the child network is copied from the fitter parent if the random probability is greater than 0.5, and vice versa.

Once a child has been created, that child will be mutated based on a predefined \textit{mutation rate}. In our work, \textit{mutation rate} is set to 0.05. A random number is generated, if that number is less than \textit{mutation rate}, mutation occurs, and vice versa. If mutation occurs for each weight in the child, that weight is added by a random number from the range [-0.05, 0.05], a slight mutation. After that, the newly born individual is placed in a new population. This process is repeated until the new population is filled 100 new individual agents. No elitism is employed in our evolutionary algorithm.

The population goes through a total of 100 generations, with 5000 time steps per generation. At each time step, an agent does the following activities: Perceiving MiniWorld through its sensors, computing its motor outputs from its sensory outputs, moving in the environment which then updates its new heading and location. In evolution alone simulation, the agent cannot perform any kind of learning during its lifetime. After that, the population undergoes selection and reproduction processes.

\subsection{Simulation 2: Evolution of Self-taught agents (EVO+Self-taught)}
In this simulation, we allow lifetime learning, in addition to the evolutionary algorithm, to update the weights of neural network controllers when agents interact with the environment. We evolve a population of \textbf{Self-taught} agents -- agents that can teach themselves. The self-taught agent has a self-taught neural network architecture as described previously and shown in Figure \ref{fig:self-taught}. During the lifetime of an agent, the reinforcement modules produce outputs in order to guide the weight-updating process of the action module. Only the weights of action modules can be changed by learning, the weights of reinforcement module are genetically specified in the same evolutionary process as specified above in Evolution alone simulation.

We can interpret this scenario as an agent has an ability to produce reinforcement signals to guide itself. It is evolution that produces these reinforcement signals, or the desire to \textit{external stimuli} (the sensory inputs in this case), for every agent. In other words, it is evolution that provides the self-teaching ability for each agent. This is how evolution influences learning. And more than this, it is learning during the lifetime that changes the fitness of each agent, hence the fitness landscape which then affects the evolutionary process. This is the interaction between learning and evolution which is being investigated in this paper.

We use the same parameter setting for evolution as in EVO simulation above. At each time step, an agent does the following activities: Perceiving MiniWorld through its sensors, computing its motor outputs from its sensory outputs, moving in the environment which then updates its new heading and location, and updating the weights in action module by \textbf{self-teaching}. After one step, the agent updates its fitness by the number of food eaten. After that, the population undergoes selection and reproduction processes as in Evolution alone.

Remember that we are fitting learning and evolution in a Darwinian framework, not Lamarckian. This means what will be learned during the lifetime of an agent (the weights in action module) is not passed down onto the offspring.

\subsection{Simulation 3: Self-taught agents alone (Self-taught-alone)}
We conduct another simulation in which all agents are self-taught agents -- having self-taught networks that can teach themselves during lifetime. What differs from simulation 2 is that at the beginning of every generation, all weights are randomly initialised, rather than updated by an evolutionary algorithm like in simulation 1. The learning agents here are initialised as \textit{blank-slates}, or \textit{tabula rasa}, having no predisposition to learn or some sort of \textit{priori knowledge} about the world. The reason for this simulation is that we are curious whether evolution brings any benefit to learning in MiniWorld. In other words, we would like to see if there is a synergy between evolution and learning, not just how learning can affect evolution.

Experimental results are presented and discussed in the following section.

\section{Results and Analysis}

\subsection{Learning Facilitates Evolution}
First we look at the performanace of the first two simulations, EVO and EVO+Self-taught. All results are averaged over 30 independent runs.

\begin{figure}[t]
\begin{center}
\includegraphics[width=5in]{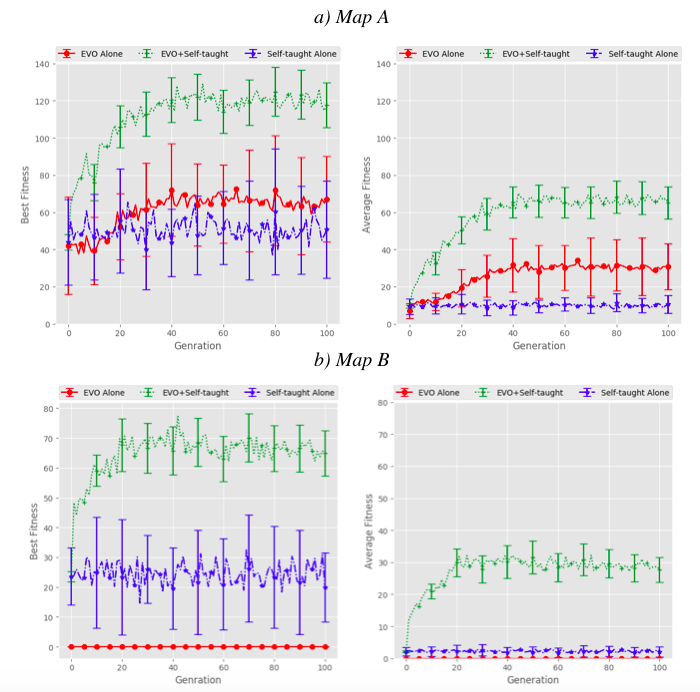}
\caption{Comparison of eating ability. a) Top: Map A b) Bottom: Map B}
\label{fig:fitness}
\end{center}
\end{figure}

\begin{figure}[t]
\begin{center}
\includegraphics[width=5in]{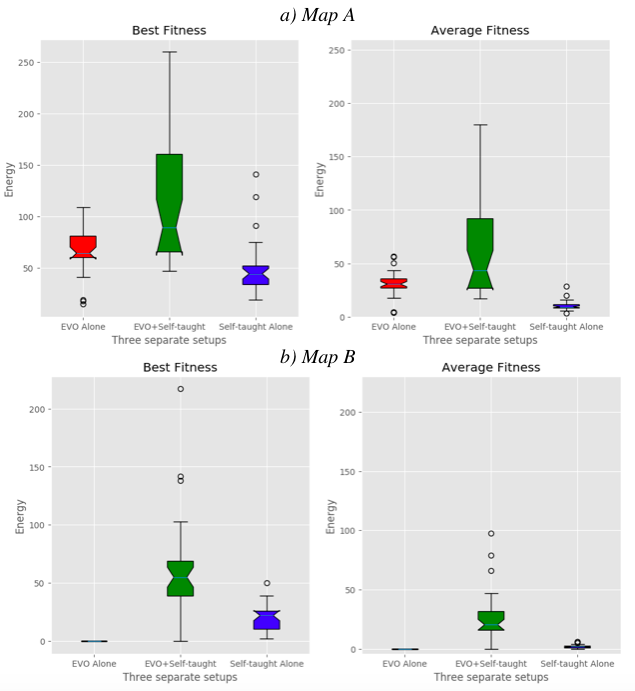}
\caption{Boxplot. a) Top: Map A b) Bottom: Map B.}
\label{fig:box-fitness}
\end{center}
\end{figure}

Figure \ref{fig:fitness} depicts the dynamics of fitness over generations, while Figure \ref{fig:box-fitness} presents a statistical comparison of the best and average fitness over runs. A similar trend can be observed is that all experimental settings have higher performance in map A than in map B. This is understandable as map B has been shown more difficult to forage than map A.

How each type of experimental setup performs compared to each other? First we look at the dynamics of the number of food eaten. It can be seen that EVO+Self-taught outperforms EVO alone in all maps with respect to both the best and average fitness. Specifically, by looking at the performance on map A we see that the best agent in EVO+Self-taught, on average, eats around 40 food particles more than the best agent in EVO alone. Additionally, the \textit{average} agent in EVO+Self-taught eats around 40 food items more than the \textit{average} agent in EVO alone. This means, as a whole, the EVO+Self-taught system has around 800 energy (each food item accounts for 1 energy) higher than the EVO system alone.

Looking at the performance on map B, it is interesting to see that while the EVO+Self-taught system still can forage for foods, the EVO system cannot eat any food at all. 

Please recall the description of our learning agents as well as the map B. Every learning agent is born with an initial horizontal heading that may be changed when the agent experiences the world through its senses and motors. The more the agent encounters, the more likely the agent can change its subsequent movements, hence its heading. However, in map B the food source is located far from the agent, at first, and far from the initial heading of every agent. This means that each agent with its innate ability and horizontal heading cannot move along the correct direction to the food source. After being born, they move horizontally as designed. Importantly, because of the inability to learn to change the behaviour during lifetime, every agent in EVO moves based on its innate behaviour. This explains why the EVO alone system cannot forage and eat food.

Conversely, the self-taught agent can still eat food in map B. One plausible explanation for this is the effect of learning through self-teaching on evolution as follows. Like in EVO alone, every self-taught agent is initially born with a wrong direction to the food source. However, with the ability to teach oneself by leveraging the difference between the action and the reinforcement modules, the weights of the action module of some agent may have been changed during lifetime by backpropagation algorithm. It is this process that may have changed the movement of some agent, make it more random at first (like performing a random search in the movement space, rather than going in one direction). By doing some random movement, there may have been some agent that somehow could reach the food source (e.g. by any kind of luck). Because of this, the agent that can reach the food source has a higher chance of being selected to produce offspring the for next generation. Thus, its genetic information is more likely to proliferate. It is important to note that the genetic information of each self-taught learning agent consists of not only the initial weights for the action module but also the initial weights for the reinforcement module. Thus, when an agent is selected for reproduction, its self-teaching ability is likely to be also promoted at later generations.

The boxplots in Figure \ref{fig:box-fitness} present some statistical results on the best and the average fitness over 30 runs. we can easily see the same effect as presented above in both map A and map B. The advantage of EVO+Self-taught over EVO alone is statistically significant.

We can claim that he combination of learning in the form of self-teaching and evolution increases the adaptivity of the population measured by the number of food eaten in any case.

\subsection{Is That the Baldwin Effect?}
We have seen how learning during lifetime facilitates the evolving population of self-taught agents, having higher performance in a multi-agent environment compared to EVO alone. One curious question here is whether the Baldwin-like Effect has occurred? 

This is why we conduct the third simulation in which the neural networks of self-taught agents are all randomly initialised, without the participation of evolution. It can be observed in Figure \ref{fig:fitness} and Figure \ref{fig:box-fitness} that in both map A and B, the population of randomly self-taught agents has lower performance than that of EVO+Self-taught, especially when it comes to the performance of the whole population (average fitness in our scenario). The difference is statistically significant as shown in Figure \ref{fig:box-fitness}.

It is also interesting that in our simulation, the \textbf{blank-slate} population by self-teaching cannot outperform the EVO alone in the easier map A, but has little advantage over the EVO alone in the harder case (map B) when the EVO alone cannot search for any food.

It is plausible here to conclude that learning, as a faster adaptation, can provide more adaptive advantage than the slower evolutionary process when the environment is dynamic like in MiniWorld. However, it is evolution that provides a good base for self-taught agents to learn better adaptive behaviours in future generations rather than learning as \textit{blank-slates} in Random-Self-taught population. This can also be explained by the understanding of the Baldwin Effect, or the synergy between evolution and learning. Through the evolutionary process, some priori-knowledge about the environment can be encoded in the neural networks controlling agents. Agents having priori-knowledge, or predisposition to learn adaptive behaviours in our scenario, can learn faster and learn more adaptively than blank-slate agents. This is the Baldwin-like Effect -- the interplay between learning and evolution.

\subsection{The Emergence of Intelligent Foraging}
Interestingly, it is not just the performance but also the emergence of foraging behaviour -- what can be called \textit{intelligent} in this sense.

\begin{figure}[t]
\begin{center}
\includegraphics[width=5in]{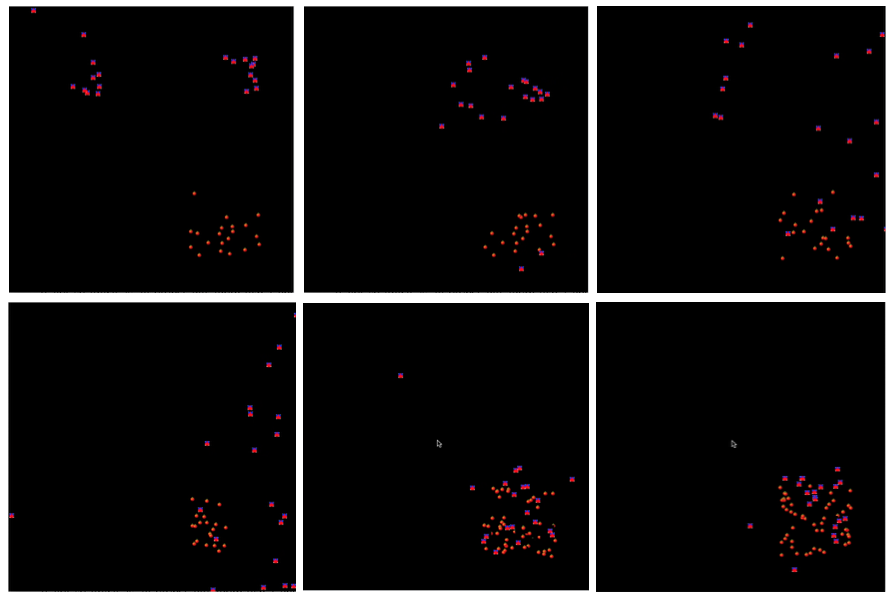}
\caption{The emergence of intelligent foraging strategy overtime by EVO+Self-taught on map B.}
\label{fig:emergence}
\end{center}
\end{figure}

Figure \ref{fig:emergence} depicts the emergence of an intelligent foraging behaviour over time. Due to scope of this paper we only report the emergence on the harder map, where only one system -- the EVO+Self-taught has shown a dominant performance. 

As we can see in the first two images in Figure \ref{fig:emergence}, at earlier generations the population cannot find the way to reach the food source. Some might have moved in some random orientation rather than just following the horizontal direction. In the two following images, we can see that after several generations some agents appeared to find a way to reach the food source, while the rest was still unable to forage correctly, moving randomly but getting better.

In the second-last image, we observe that most agents have found the way to reach the food source except for one agent. However, the last image shows the whole system could reach the food source. They stayed there and competed for resources. All agents seem to know where to forage as a whole. Remember that, every agent in our MiniWorld does not have any idea about the location of the map (very simple sensory inputs) as well as the location of other agents. However, at the end of the day they still can reach the food source. This intelligence is the emergence through the interaction between evolution and self-teaching in the evolution of their brains, or neural networks.

\section{Conclusion and Future Work}
In this paper, we have presented a technique called Evolving Self-taught Neural Networks, and simulated a foraging task in a multi-agent system. Experimental results have shown that the proposed technique which combines evolutionary search and self-teaching in neural networks can enhance the system, better than evolution and self-teaching staying in isolation. An intelligent foraging behaviour is shown to emerge from the interaction between evolution and self-teaching. Self-teaching ability can help an agent better adapt to its environment, changing the subsequent evolutionary pathway of a species. Evolution is shown to provide more adaptive self-taught agents in future generations, better than learning as \textit{blank-slates}.

There are quite a few avenues for future research building-on this study. We can complexify MiniWorld by including more objects like obstacles, more substances with negative rewards (poison) to make the learning task more complex, hence the intelligence required. We are curious to see how the evolved self-teaching ability can better promote the system in more complex environments.

The computational method is simple enough to illustrate the idea, but still has some indications. The idea of self-taught neural networks can be powerful when there is no external supervision (or \textit{label} provided from external data). This opens a way to produce autonomous intelligence, which might open a route to AGI -- Artificial General Intelligence. The algorithm and technique used in this paper can also be a potential technique to solve unsupervised learning, or learning with limited label data (weak supervision, especially in reinforcement learning and games.  We are curious whether evolution can provide a better base to learn than learning as blank-slates like what was claimed by DeepMind in games \cite{Mnih2015}. Indeed, the shallow network used in this paper does not restrict the application of the core philosophical idea into deep neural networks, as long as we can combine evolutionary search and the idea of self-taught neural architecture by employing variants of gradient-based learning.

There is some limitation that should not be neglected, including the use of a fixed neural architecture. One plausible solution could be evolving both the weights and the topology of a neural network \cite{stanley:cec03}. This is an interesting pathway for future work if we can evolve variable self-supervised neural architecture which can be an intrinsically general neural learner.

Delving a little deeper into lifetime learning, this category can be subdivided into \textit{asocial} (or individual) learning (IL) and \textit{social} learning (SL). Each is a plausible way for an individual agent to acquire information from the environment at the phenotypic level. SL has been observed in organisms as diverse as primates, birds, fruit flies, and especially humans \cite{rendell:2011}. Self-teaching can be considered an individual learning process which updates the behaviour of a single agent. The relationship between individual and social learning has raised some important scientific curiosity as whether the organism should rely on social or individual information \cite{Laland:2004}, \cite{LeNam:PPSN:2018}. Social learning may offer another way to propose where the reinforcement signal comes from. If it learns from observing other agents, then the self-learning could proceed from imitation learning. Future work will investigate this line of research and see if the presence of social learning could result in a more complex intelligent behaviour.

\bibliographystyle{unsrt}  


\end{document}